%% file: main.tex
\documentclass[10pt,twocolumn,letterpaper]{article}

\usepackage{cvpr}      % To produce the REVIEW version
\usepackage{newfloat}
\usepackage{listings}

\usepackage{amsmath}
\newtheorem{definition}{Definition}
\newtheorem{lemma}{Lemma}

\usepackage{multirow}
\usepackage{amsfonts}       % blackboard math symbols
\usepackage{makecell}

\definecolor{cvprblue}{rgb}{0.21,0.49,0.74}
\usepackage[pagebackref,breaklinks,colorlinks,allcolors=cvprblue]{hyperref}

\def\paperID{*****} % *** Enter the Paper ID here
\def\confName{CVPR}
\def\confYear{2025}

\title{Seeking the Necessary and Sufficient Causal Features\\ in Multimodal Representation Learning}
\author{
    Boyu Chen\textsuperscript{\rm 1}, Junjie Liu\textsuperscript{\rm 2}, Zhu Li\textsuperscript{\rm 3}, Mengyue Yang\textsuperscript{\rm 4,5}\thanks{Corresponding author}\\
    \small \textsuperscript{\rm 1}Institute of Health Informatics, University College London\\
    \small \textsuperscript{\rm 2}Department of Aeronautics, Imperial College London\\
    \small \textsuperscript{\rm 3}Gatsby Computational Neuroscience Unit, University College London\\
    \small \textsuperscript{\rm 4}Department of Computer Science, University College London\\
    \small \textsuperscript{\rm 5}School of Engineering Mathematics and Technology, University of Bristol\\
}

\begin{document}
\maketitle
\begin{abstract}
Probability of necessity and sufficiency (PNS) measures the likelihood of a feature set being both necessary and sufficient for predicting an outcome. It has proven effective in guiding representation learning for unimodal data, enhancing both predictive performance and model robustness. Despite these benefits, extending PNS to multimodal settings remains unexplored. This extension presents unique challenges, as the conditions for PNS estimation—exogeneity and monotonicity—need to be reconsidered in a multimodal context. We address these challenges by first conceptualizing multimodal representations as comprising modality-invariant and modality-specific components. We then analyze how to compute PNS for each component while ensuring non-trivial PNS estimation. Based on these analyses, we formulate tractable optimization objectives that enable multimodal models to learn high-PNS representations. Experiments demonstrate the effectiveness of our method on both synthetic and real-world data.
\end{abstract}

\section{Introduction}
Probability of necessity and sufficiency (PNS) measures the likelihood of a feature set being both necessary (without which the outcome cannot occur) and sufficient (which guarantees the outcome) for an outcome \cite{pearl2009causality}. Recent studies have shown that learning high-PNS representations can enhance both the predictive performance and robustness of models trained on unimodal data \cite{yang2024invariant, wang2021desiderata, chen2024unifying, cai2024learning, chen2024medical}. Despite these benefits and the increasing importance of learning meaningful representations from diverse modalities \cite{xu2024multimodal, liu2024muse, liang2024module, liang2024quantifying, swamy2024multimodn, tang2024brain, dong2023dreamllm}, extending PNS to multimodal contexts remains underexplored. Such an extension represents a promising direction for multimodal models. It has the potential to improve both their predictive capabilities through better feature capture and enhance their robustness under missing modalities.

Nevertheless, this extension faces challenges in satisfying two conditions for PNS estimation: exogeneity and monotonicity. Exogeneity requires causal features to be determined independently of unmeasured confounders and the system's internal dynamics. In multimodal scenarios, inter-modal interactions can compromise this condition. Moreover, treating multimodal data as unimodal can violate exogeneity, since cross-modal dependencies can introduce hidden confounding effects that are difficult to isolate without strong assumptions or additional supervision. \cite{yang2024invariant, wang2021desiderata, locatello2019challenging, liu2021learning}. On the other hand, monotonicity requires causal features to monotonically influence outcome prediction. However, the complex interactions in multimodal data often result in non-monotonic relationships, and their high-dimensional nature complicates the assessment of consistent directional effects across modalities.

To address these challenges, instead of analyzing PNS on the whole multimodal representation, we propose viewing the representation as two parts: a modality-invariant component that captures information shared across modalities, and modality-specific components that preserve the unique characteristics of each modality \cite{zhang2019multimodal, ramachandram2017deep, guo2019deep, gao2020survey, li2023decoupled}. This decomposition enables separate analysis of each component, making it possible to establish tractable conditions for non-trivial PNS estimation. With these insights, we develop optimization objectives for learning high-PNS multimodal representations.

Our main contributions are: (1) introducing PNS in multimodal representation learning and analyzing its challenges, (2) proposing to consider multimodal features as two components and derive PNS estimation tailored for these components, (3) developing optimization objectives based on these findings to enhance multimodal learning. Experimental results on both synthetic and real-world datasets demonstrate the effectiveness of our method.

\section{Related Works}
\textbf{Causal representation learning. }
Causal representation learning aims to identify underlying causal information from observational data, enhancing machine learning models' trustworthiness through improved explanation, generalization, and robustness \cite{arjovsky2019invariant, hu2018causal, ahuja2020invariant, gamella2020active, scholkopf2021toward}. This field encompasses two directions: causal relationship discovery \cite{peters2014causal, zheng2018dags, huang2020causal, zhu2019causal}, which uncovers causal structure among variables, and causal feature learning \cite{zhang2020invariant, chen2022learning, louizos2017causal, yang2021learning, lu2021invariant}, which extracts features that causally influence the target outcome.
Recently, PNS has emerged as a powerful tool for causal feature learning and has demonstrated success in improving deep learning model performance. Its applications include learning invariant representations for out-of-distribution generalization \cite{yang2024invariant}, identifying crucial genes \cite{cai2024learning}, formulating efficient low-dimensional representations \cite{wang2021desiderata}, and improving medical image quality assessment \cite{chen2024medical}. However, these applications primarily focus on unimodal data, leaving multimodal scenarios unexplored.

\textbf{Multimodal Representation Decomposition.}
Multimodal learning captures meaningful representations from multiple modalities \cite{islam2023eqa, zhang2023integration, tran2024training, liu2024focal}. Among various approaches, a family of models that decompose multimodal representations \cite{shi2019variational, tsai2018learning, mai2020modality, wang2017adversarial, li2023decoupled} has emerged as a promising direction. These models, which we refer to as ``decomposition models", decouple the representations into two components: a modality-invariant component that captures shared semantic information, and modality-specific component that preserves unique characteristics within each modality.

Our work bridges these research areas by extending PNS estimation to multimodal settings through a novel decomposition perspective. Specifically, we analyze PNS computation by viewing multimodal representations as two components. This  not only simplifies PNS analysis but also enables the utilization of existing decomposition models to extract these components.

\begin{figure}[b]
\centerline{\includegraphics[width=5cm]{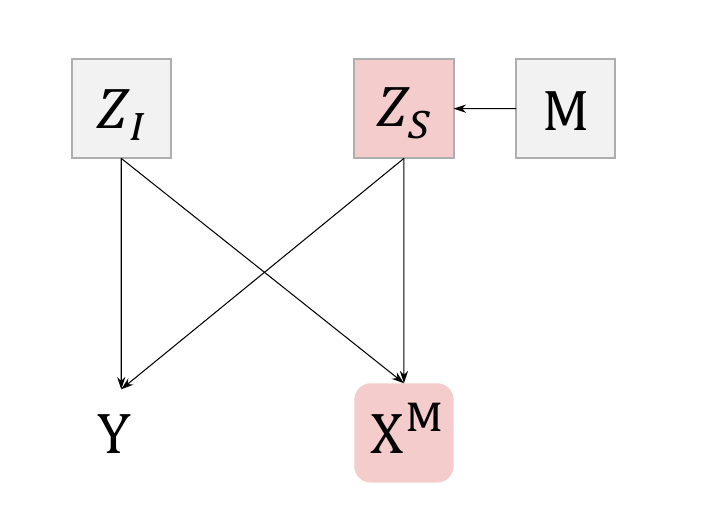}}
\caption{The causal graph showing data generation process with modality $M$}
\label{fig:data_generation_process}
\end{figure}

\section{Preliminaries}
\subsection{Problem Setup}
\label{subsec:Preliminaries_setup}
Let $(X^M, Y)$ denote a multimodal variable of modality $M$, where $X^M \subset \mathbb{R}^{d_M}$ represents the features and $Y \subset \mathbb{R}^{d_y}$ represents the labels, with dimensionalities $d_M$ and $d_y$ respectively. For a set of $N$ modalities, we use $m \in \{1, \ldots, N\}$ to represent specific modalities. A unimodal sample from modality $m$ is denoted as $(X^m, Y)$, where a multimodal sample $(X, Y)$ consists of samples from all modalities, written as $(\{X^M\}_{M=1}^{N}, Y)$, with its specific instance denoted as $(\{x^M\}_{M=1}^{N}, Y)$.

Following prior work \cite{shi2019variational, tsai2018learning, mai2020modality, wang2017adversarial, li2023decoupled}, $(X^M, Y)$ can be decomposed into modality-invariant and modality-specific hidden features. The data generation process is illustrated in \cref{fig:data_generation_process}, which involves two key latent variables: a modality-invariant variable $Z_I \subset \mathbb{R}^{d_{Z_I}}$ that captures cross-modal shared information, and a modality-specific variable $Z_S \subset \mathbb{R}^{d_{Z_S}}$ that encodes unique characteristics conditioned on $M$, where $d_{Z_I}$ and $d_{Z_S}$ denote their respective dimensionalities.

\subsection{Probability of Necessity and Sufficiency (PNS)}
\label{sec:PNS_intro}
The PNS measures the likelihood of a feature set being both necessary and sufficient for an outcome. A feature is considered necessary if it is indispensable for causing an outcome, and sufficient if it alone can ensure the outcome.

\begin{definition}[PNS \cite{pearl2009causality}]
\label{def:pns}
Let $Z$ be the causal features of outcome $Y$, with $z$ and $\bar{z}$ being two distinct values of $Z$. The PNS of $Z$ with respect to $Y$ for $z$ and $\bar{z}$ is defined as:
\begin{equation} \nonumber
\begin{aligned}
&\text{PNS}(z, \bar{z}) := \\
&P(Y_{\mathrm{do}(Z=z)}=y|Z=\bar{z},Y \neq y)P(Z=\bar{z},Y \neq y) \\
&+ P(Y_{\mathrm{do}(Z=\bar{z})}\neq y|Z=z,Y=y)P(Z=z,Y=y)\\
\end{aligned}
\end{equation}
\end{definition}
Here, $P(Y_{\mathrm{do}(Z=z)}=y|Z=\bar{z},Y \neq y)$ represents the counterfactual probability that $Y=y$ when $Z$ is set to $z$ (via the do-operator), given the factual observation $Z=\bar{z}$ and $Y \neq y$. An analogous interpretation holds for the counterfactual probability $P(Y_{\mathrm{do}(Z=\bar{z})}\neq y|Z=z,Y=y)$. Although a high PNS indicates stronger necessity and sufficiency of $Z$ for $Y$, computing counterfactual probabilities is difficult due to the challenges in obtaining counterfactual data. However, under exogeneity and monotonicity conditions, PNS can be estimated from observational data.

\begin{definition}[Exogeneity \cite{pearl2009causality}]
$Z$ is exogenous to $Y$ if the intervention probability can be expressed as a conditional probability: $P(Y_{do(Z=z)} = y) = P(Y = y \mid Z=z)$.
\end{definition}
\begin{definition}[Monotonicity \cite{pearl2009causality}]
$Y$ is monotonic with respect to $Z$ if and only if either $(Y_{do(Z=z)} \neq y) \wedge (Y_{do(Z=\bar{z})} = y)$ is false, or $(Y_{do(Z=z)} = y) \wedge (Y_{do(Z=\bar{z})} \neq y)$ is false. This can be presented as: $P(Y_{do(Z=z)} = y)P(Y_{do(Z=\bar{z})} \neq y) = 0$ or $P(Y_{do(Z=z)} \neq y)P(Y_{do(Z=\bar{z})} = y) = 0$.
\label{def:mono}
\end{definition}
\begin{lemma}[\cite{pearl2009causality}]
If $Y$ is monotonic relative to $Z$, then:
\begin{equation} \nonumber
\label{eq:pns_identifiability_mono}
\begin{aligned}
\text{PNS}(z, \bar{z}) := & P(Y_{\mathrm{do}(Z=z)}=y) - P(Y_{\mathrm{do}(Z=\bar{z})}=y)
\end{aligned}
\end{equation}
\end{lemma}

\begin{lemma}[\cite{pearl2009causality}]
\label{lemma:pns}
If $Z$ is exogenous relative to $Y$, and $Y$ is monotonic relative to $Z$, then:
\begin{equation} \nonumber
\label{eq:pns_identifiability_exo}
\begin{aligned}
\text{PNS}(z, \bar{z}) := & P(Y = y \mid Z = z) - P(Y = y \mid Z = \bar{z})
\end{aligned}
\end{equation}
\end{lemma}
\cref{lemma:pns} enables PNS computation using real-world data when counterfactual data is unavailable, provided both exogeneity and monotonicity conditions hold.

\begin{figure*}[t]
\centerline{\includegraphics[width=0.9\textwidth]{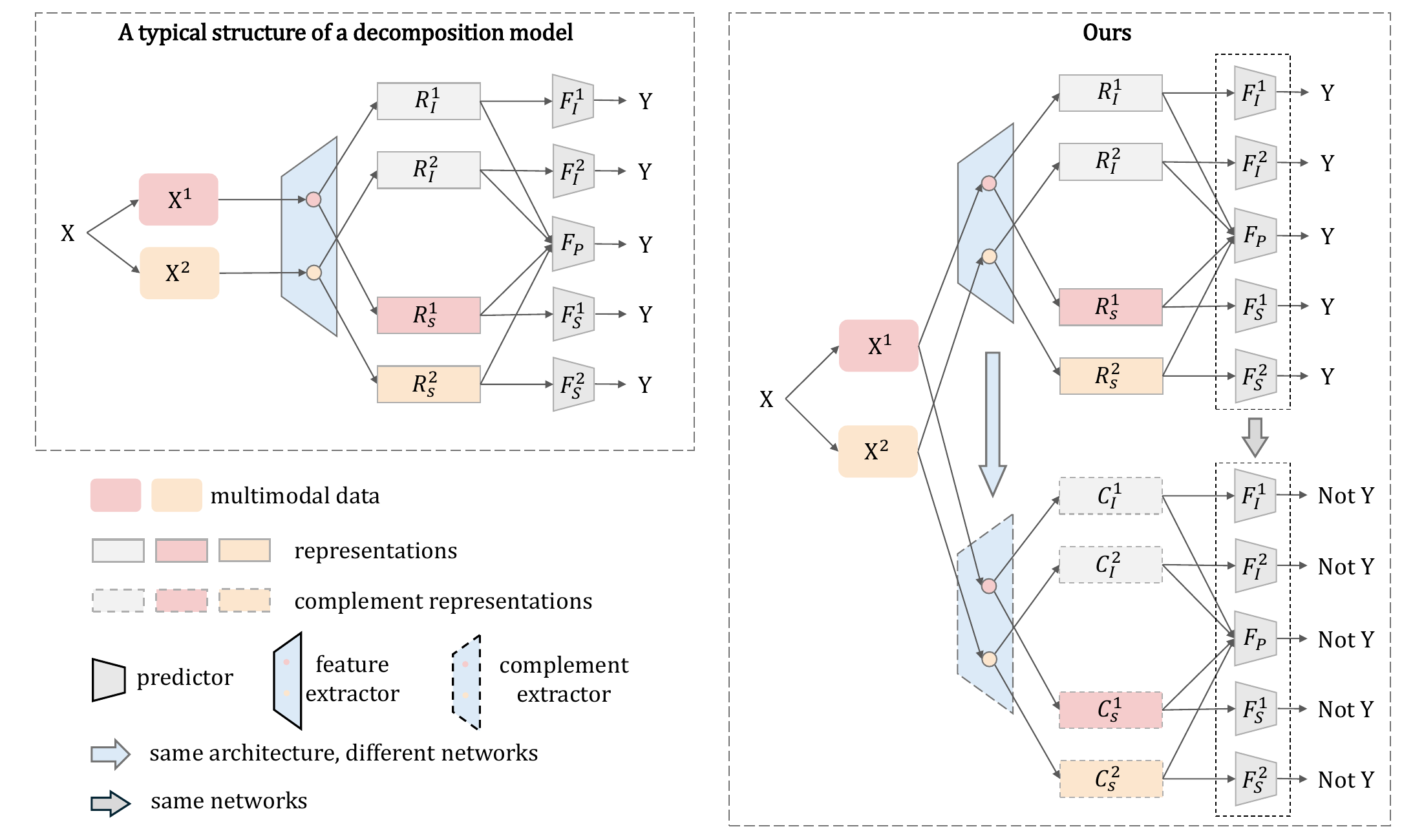}}
\caption{A typical structure of a decomposition model and its adaptation to our method}
\label{fig:method}
\end{figure*}

\section{PNS in Multimodality}
\label{sec:pns4multimodal}
We view multimodal data as being composed of modality-invariant ($Z_I$) and modality-specific ($Z_S$) hidden variables (\cref{fig:data_generation_process}). This section analyzes how to compute PNS for them, which forms the foundation for designing optimization objectives in the next section.

\subsection{PNS for Modality-Invariant Variables}
If $Z_{I}$ satisfies monotonicity, we can estimate its PNS by:
\begin{equation} \nonumber
\label{eq:pns_mi}
\begin{aligned}
\text{PNS}_I(z, \bar{z}) :=& P(Y_{\mathrm{do}(Z_I=z)}=y) - P(Y_{\mathrm{do}(Z_I=\bar{z})}=y)
\end{aligned}
\end{equation}

As illustrated in \cref{fig:data_generation_process}, $Z_I$ naturally satisfies exogeneity since it directly influences $Y$ without being affected by other variables, leading to $P(Y_{\mathrm{do}(Z_I)}=y) = P(Y=y \mid Z_I)$. Consequently, the PNS for modality-invariant features can be computed directly from observational data:
\begin{equation}
\label{eq:pns_mi_cond}
\begin{aligned}
\text{PNS}_I(z, \bar{z}) := & P(Y=y \mid Z_I=z) \\
& - P(Y=y \mid Z_I=\bar{z})
\end{aligned}
\end{equation}

This suggests that, under monotonicity, the PNS of $Z_I$ can be estimate based on observational data. Constraints for monotonicity are designed during the learning process, which will be discussed in the next section.

\subsection{PNS for Modality-Specific Variables}
Computing PNS for $Z_S$ presents unique challenges compared to $Z_I$. Under the monotonicity, the PNS for $Z_S$ is expressed as:
\begin{equation}
\label{eq:pns_ms}
\begin{aligned}
\text{PNS}_S(z, \bar{z}) :=& P(Y_{\mathrm{do}(Z_S=z)}=y) \\
& - P(Y_{\mathrm{do}(Z_S=\bar{z})}=y)
\end{aligned}
\end{equation}

However, unlike $Z_I$, the exogeneity does not hold for $Z_S$ as it is conditioned by modality type $M$ (see \cref{fig:data_generation_process}). This means $P(Y_{\mathrm{do}(Z_S)}=y) \neq P(Y=y \mid Z_S)$ and we cannot directly apply \cref{lemma:pns} to estimate the PNS using observational data. Nevertheless, we can develop an alternative estimation approach by exploiting the inherent properties of multimodal data. Consider a multimodal instance $(\{x^M\}_{M=1}^{N}, y)$, where different modalities share the same label $y$. For any two distinct modalities $m$ and $\bar{m}$, their $\text{PNS}_M(m, \bar{m})$ must be zero as they lead to the same outcome:
\begin{equation}
\label{eq:pns_m_do}
\begin{aligned}
\text{PNS}_M(m, \bar{m}) := & P(Y_{\mathrm{do}(M=m)}=y) \\
&- P(Y_{\mathrm{do}(M=\bar{m})}=y))\\
= & 0
\end{aligned}
\end{equation}
Using the front-door criterion (\cref{fig:data_generation_process}), we can decompose the intervention probabilities as:
\begin{equation}\nonumber
\begin{aligned}
&P(Y_{\mathrm{do}(M=m)}=y) := \\
&\int_{z} P(Y_{\mathrm{do}(Z_S=z)}=y)P({Z_S}_{\mathrm{do}(M=m)}=z) dz\\
\end{aligned}
\end{equation}
and
\begin{equation}\nonumber
\begin{aligned}
&P(Y_{\mathrm{do}(M=\bar{m})}=y) := \\
&\int_{\bar{z}} P(Y_{\mathrm{do}(Z_S=\bar{z})}=y)P({Z_S}_{\mathrm{do}(M=\bar{m})}=\bar{z}) d\bar{z}
\end{aligned}
\end{equation}
Substituting these into \cref{eq:pns_m_do} gives:
\begin{equation}
%\label{eq:pns}
\begin{aligned} \nonumber
\label{eq:pns_m_zero}
& \text{PNS}_M(m, \bar{m}) = \\
&\int_{z} P(Y_{\mathrm{do}(Z_S=z)}=y)P({Z_S}_{\mathrm{do}(M=m)}=z) dz\\
& - \int_{\bar{z}} P(Y_{\mathrm{do}(Z_S=\bar{z})}=y)P({Z_S}_{\mathrm{do}(M=\bar{m})}=\bar{z}) d\bar{z} \\
& =  0
\end{aligned}
\end{equation}

For an instance $(\{x^M\}_{M=1}^{N}, y)$, modality $m$ and $\bar{m}$ correspond to unique modality-specific hidden states $z$ and $\bar{z}$, respectively. This yields:
\begin{equation}
\label{eq:non_zero_eq}
\begin{aligned}
& \text{PNS}_M(m, \bar{m}) = \\
&P(Y_{\mathrm{do}(Z_S=z)}=y)P({Z_S}_{\mathrm{do}(M=m)}=z)\\
& - P(Y_{\mathrm{do}(Z_S=\bar{z})}=y)P({Z_S}_{\mathrm{do}(M=\bar{m})}=\bar{z}) \\
& =  0
\end{aligned}
\end{equation}

The terms $P(Y_{\mathrm{do}(Z_S=z)}=y)$ and $P({Z_S}_{\mathrm{do}(M=m)}=z)$ can be interpreted as predictor and feature inference components, respectively. Based on \cref{eq:non_zero_eq}, to ensure non-zero \cref{eq:pns_ms}, we must ensure:
\begin{equation} \nonumber
P({Z_S}_{\mathrm{do}(M=m)}=z) \neq P({Z_S}_{\mathrm{do}(M=\bar{m})}=\bar{z})
\end{equation}
which can be translated to learn the mapping $\mathcal{F}: \mathbb{R}^{Z_S} \rightarrow \mathbb{R}^{dy}$ that that selects features ensuring:
\begin{equation}
\label{eq:mapping}
P(\mathcal{F}(z|m) \neq \mathcal{F}(\bar{z}|\bar{m})) > const
\end{equation}
where $const$ is a positive constant. A high value of \cref{eq:mapping} serves two purposes: (1) it enforces monotonicity by ensuring changes in $Z_S$ lead to changes in predicted $Y$, and (2) it guarantees a non-trivial $\text{PNS}_S(z, \bar{z})$. In the next section, we will introduce learning constraints to satisfy this condition.

\section{Multimodal Learning via PNS}
For a specific modality $X^m$, representations with high PNS values contain both necessary and sufficient causal information for prediction. This section presents our approach to learning such representations in multimodal scenarios. Based on the analysis in \cref{sec:pns4multimodal}, we decompose $X^m$ into modality-invariant and modality-specific parts, and design specific objectives to optimize their PNS values.

\subsection{Decomposing Multimodal Features}
The foundation of our approach builds upon decomposition models \cite{shi2019variational, tsai2018learning, mai2020modality, wang2017adversarial, li2023decoupled}, which extract modality-invariant and modality-specific features from multimodal data. \cref{fig:method} (left-top) illustrates a typical structure of these models, consisting of a feature extractor and associated predictors.

The feature extractor $\Phi(\cdot)$ decomposes input $X^m$ into a modality-invariant representation $\mathcal{R}_I^m \subset \mathbb{R}^{d_{Z_I}}$ and a modality-specific representation $\mathcal{R}_S^m \subset \mathbb{R}^{d_{Z_S}}$, aiming to capture the underlying latent variables $Z_I$ and $Z_S$, respectively. This is denoted as $[\mathcal{R}_{I}^{m}, \mathcal{R}_{S}^{m}]:=\Phi(X^m)$.

For prediction, the main predictor $F_P(\cdot)$ uses the complete set of representations $[\mathcal{R}_{I}^{1}, \mathcal{R}_{S}^{1}, \mathcal{R}_{I}^{2}, \mathcal{R}_{S}^{2}, \ldots, \mathcal{R}_{I}^{N}, \mathcal{R}_{S}^{N}]$ to predict $Y$. Additionally, auxiliary predictors $F_{I}^{m}(\cdot)$ and $F_{S}^{m}(\cdot)$ are employed during training to predict $Y$ based on $\mathcal{R}_{I}^{m}$ and $\mathcal{R}_{S}^{m}$, respectively, to ensure the representations capture outcome-related information.

To compute PNS, we need the complement $\bar{z}$ for feature value $z$ of $Z$. This means finding complement modality-invariant representation $\mathcal{C}_I^m \subset \mathbb{R}^{d_{Z_I}}$ for $\mathcal{R}_{I}^{m}$ and complement modality-specific representation $\mathcal{C}_S^m \subset \mathbb{R}^{d_{Z_S}}$ for $\mathcal{R}_{S}^{m}$. Both $\mathcal{C}_{I}^{m}$ and $\mathcal{C}_{S}^{m}$ should maintain similar properties to $\mathcal{R}_{I}^{m}$ and $\mathcal{R}_{S}^{m}$, respectively, while leading to different outcome predictions. For instance, if $F_{I}^{m}(\mathcal{R}_{I}^{m})$ predicts $Y$, then $F_{I}^{m}(\mathcal{C}_{I}^{m})$ should predict a label different from $Y$.

Given the challenge of directly obtaining complement representations in real-world settings, we propose using an complement extractor $\phi(\cdot)$, shown as right-bottom in \cref{fig:method}. $\phi(\cdot)$ shares the same structure as $\Phi(\cdot)$ but is a separate network. It can learn the complement representations for $X^m$ as: $[\mathcal{C}_{I}^{m}, \mathcal{C}_{S}^{m}]:=\phi(X^m)$.

In the training process, $\phi(\cdot)$ extracts $\mathcal{C}_{I}^{m}$ and $\mathcal{C}_{S}^{m}$ from $X^m$, and the auxiliary predictors use them to predict outcomes that differ from $Y$. The rationale is to extract the complement features through a process analogous to the original feature extraction, preserving the underlying data structure while introducing meaningful variations.

By integrating $\Phi(\cdot)$, $\phi(\cdot)$, and predictors, we establish a new decomposition framework (right part of \cref{fig:method}). This framework, guided by our analysis in \cref{sec:pns4multimodal}, enables us to design specific objectives for decomposed representations. 

For optimization purposes, we define two types of loss functions: $\mathcal{L}_{p}(Y, \hat{Y})$ which decreases as predicted label $\hat{Y}$ approaches ground truth $Y$, and $\mathcal{L}_{c}(Y, \hat{Y})$ which decreases as $\hat{Y}$ deviates from $Y$. Their specific implementations depend on the task and will be detailed in the experiments.

To ensure interpretability in PNS calculations, we adopt the widely accepted assumption of semantic separability: small changes in representations can preserve their semantic meaning \cite{yang2024invariant}. Specifically, features extracted by different extractors from the same input maintain their respective semantic meaning.

\subsection{PNS for Modality-Invariant Representation}
We design the following objective to encourage learning high-PNS modality-invariant features from $X^m$:

\begin{equation}
\label{eq:pns_I_loss}
\begin{aligned}
& \mathcal{L}_{m,I}^{pns} :=  \mathcal{L}_{m,I}^{r} + \mathcal{L}_{m,I}^{cr} + \mathcal{L}_{m,I}^{constr} 
\end{aligned}
\end{equation}

The $\mathcal{L}_{m,I}^{r}$ is defined as $\mathcal{L}_{p}(Y, F_{I}^{m}(\mathcal{R}_I^m)) $. Optimizing this term increases the probability of the prediction being close to $Y$ when the modality-invariant representation is set to $\mathcal{R}_I^m$. This aims to encourage representation to capture a high $P(Y=y \mid Z_I=z)$ in \cref{eq:pns_mi}. 

The $\mathcal{L}_{m,I}^{cr}$ is defined as $\mathcal{L}_{c}(Y, F_{I}^{m}(\mathcal{C}_I^m))$. Optimizing this term decreases the probability of the prediction being close to $Y$ when the modality-invariant representation is set to $\mathcal{C}_I^m$. This helps learn the representations that capture a low value for $P(Y=y \mid Z_I=\bar{z})$ in \cref{eq:pns_mi}. 

Together, optimizing $\mathcal{L}_{m,I}^{r} + \mathcal{L}_{m,I}^{cr}$ represents the process of improving the PNS in \cref{eq:pns_mi}.

The $\mathcal{L}_{m,I}^{constr}$ serves as a monotonicity constraint and is defined as $\mathcal{L}_{p}(Y, F_{I}^{m}(\mathcal{R}_I^m)) * \mathcal{L}_{c}(Y, F_{I}^{m}(\mathcal{C}_I^m))$. Optimizing this term encourages representations to satisfy $P(Y_{do(Z=z)} \neq y)P(Y_{do(Z=\bar{z})} = y)=0$ in \cref{def:mono}, as the multiplication of probabilities decreases when $\mathcal{L}_{m,I}^{constr}$ decreases. This aims to foster an environment where the monotonicity is more likely to be met.

\subsection{PNS for Modality-Specific Representation}
We design the following objective to encourage learning high-PNS modality-specific features from $X^m$:
\begin{equation}
\mathcal{L}_{m,S}^{pns} :=  \mathcal{L}_{m,S}^{r} + \mathcal{L}_{m,S}^{cr} + \mathcal{L}_{m,S}^{constr}
\label{eq:pns_S_loss}
\end{equation}

The $\mathcal{L}_{m,S}^{r}$ and $\mathcal{L}_{m,S}^{cr}$ are defined as $\mathcal{L}_{p}(Y, F_S^m(\mathcal{R}_S^m))$ and $\mathcal{L}_{c}(Y, F_S^m(\mathcal{C}_S^m))$ to learn $\mathcal{R}_S^m$ and $\mathcal{C}_S^m$, respectively.

We design the constraint term $\mathcal{L}_{m,S}^{constr} = \mathcal{L}_{c}(F_S^m(\mathcal{R}_S^m), F_S^{m}(\mathcal{C}_S^{\bar{m}}))$, where $m \neq \bar{m}$. Optimizing this term aims to increase $P(\mathcal{F}(z|m) \neq \mathcal{F}(\bar{z}|\bar{m}))$ in \cref{eq:mapping} as this probability increase when $F_S^m(\mathcal{R}_S^m$) deviates from $F_S^{m}(\mathcal{C}_S^{\bar{m}})$, thereby facilitating non-trivial PNS estimation. Here, $\mathcal{C}_S^{\bar{m}}$ is generated by $\phi(X^{\bar{m}})$, where $X^m$ and $X^{\bar{m}}$ are different modalities of the same multimodal $X$.

\subsection{Multimodal PNS Learning }
We design the following objective to encourage learning high-PNS representation from multimodal sample $(X, Y)$:
\begin{equation}
\label{eq:pns_loss}
\begin{aligned}
\mathcal{L}^{total} := & \mathcal{L}^{task} + \sum_{M=1}^{N}( \mathcal{L}_{M,I}^{pns} + \mathcal{L}_{M,S}^{pns}) % \\
% = & \mathcal{L}^{task} + \mathcal{L}^{pns} + \mathcal{L}^{constr}\\
\end{aligned}
\end{equation}
where $\mathcal{L}^{task}$ is the original loss of the decomposition model.

We name our approach MPNS (\underline{M}ultimodal Representation Learning via \underline{PNS}). Implementing MPNS is straightforward: First, select a base decomposition model (left-top of \cref{fig:method}). Second, construct its complement feature extractor to form an enhanced decomposition framework (right of \cref{fig:method}), where training incorporates PNS-oriented objectives $\sum_{M=1}^{N}( \mathcal{L}_{M,I}^{pns} + \mathcal{L}_{M,S}^{pns})$ alongside the original loss $\mathcal{L}^{task}$. Once training completes, the complement extractor and auxiliary predictors are discarded, leaving only the original base model for inference.

\section{Experiment}
\label{sec:experiment}
We evaluate MPNS using both synthetic and real-world datasets. First, we construct a synthetic dataset to show that MPNS can capture high-PNS representations. Then, we utilize real-world datasets to demonstrate MPNS's ability to enhance the predictive performance and robustness of its adapted decomposition model. All experiments are conducted on a Linux system with an NVIDIA Tesla V100 PCIe GPU.

\subsection{Synthetic Dataset Experiments}
We construct a synthetic dataset to demonstrate MPNS's effectiveness in learning essential information (necessary and sufficient causes) from multimodal data. We adapt the data generation and evaluation process from \cite{yang2024invariant}. This process involves generating deterministic variables that directly determine the outcome, along with other variables, which are then mixed. Subsequently, representations are extracted from these mixed variables by a neural network to predict outcomes. For evaluation, we use Distance Correlation \cite{jones1995fitness} to measure how well each type of variable is captured in the learned representations. Higher correlation values indicate more relevant information captured. As deterministic variables directly influence the outcome, they possess high PNS. Consequently, a method achieving high distance correlation between deterministic variables and representations can effectively captures essential, high-PNS information \cite{yang2024invariant}.

\subsubsection{Generating the Synthetic Dataset.} We generate a synthetic dataset based on four types of variables. These variables are used to construct a two-modality sample and its corresponding label:

\textbf{Sufficient and Necessary (SN) cause variable $sn$} is the deterministic variable and generated from a Bernoulli distribution $B(0.5)$, with probability of 0.5 to be 1. It directly determines the label $Y$ through the relationship $Y = sn \oplus B(0.15)$, where $\oplus$ is the XOR operation.

\textbf{Sufficient and Unnecessary (SF) cause variable $sf$} is generated by transforming $sn$. When $sn = 0$, $sf = B(0.1)$, and when $sn = 1$, $sf = sn$.

\textbf{Insufficient and Necessary (NC) cause variable $nc$} is generated as $I(sn = 1) \cdot B(0.9)$, where $I(\cdot)$ is indicator function.

\textbf{Spurious correlation (SC) variable $sc$} is generated to have a spurious correlation with the SN cause, defined as $s \cdot sn + (1 - s) \mathcal{N}(0, 1)$, where $s \in [0,1)$ is the degree of spurious correlation and $\mathcal{N}(0, 1)$ denotes the standard Gaussian distribution. 

Based on these variables, we construct a feature vector $h = [sn \cdot \mathbf{1}_d, sf \cdot \mathbf{1}_d, nc \cdot \mathbf{1}_d, sp \cdot \mathbf{1}_d] + \mathcal{N}(0, 0.3)$, where $\mathbf{1}_d$ is a $d$-dimensional vector of ones and $d$ is set to 7. Following \cref{fig:data_generation_process}, we create synthetic multimodal data with modality-invariant and modality-specific components. The first 3 elements of each variable serve as the modality-invariant component. For modality-specific features, we allocate the next 2 elements to modality 1 and the last 2 to modality 2. We then form temporary feature vectors $h^1$ and $h^2$ for each modality by combining the invariant component with their respective specific elements. To introduce varying complexities between two modalities, we apply a non-linear function $\kappa(t, \alpha, \beta) = \beta \cdot \max(t - \alpha, 0) \cdot \min(t + \alpha, 0)$. The final multimodal sample $[X^1, X^2, Y]$ is generated as $X^1 = \kappa(h^1, 0.8, 2.2)$ and $X^2 = \kappa(h^2, 1, 2)$. 

To analyze the impact of different levels of spurious correlation on the learned representations, we vary the $s$ as 0.0, 0.1, 0.3, 0.5, and 0.7. For each value of s, we generate 15,000 samples for training and 5,000 for evaluation.

\subsubsection{decomposition model.}
We refer to \cite{li2023decoupled} to design a simple decomposition model. Specifically, we construct feature extractor by exploiting a shared multimodal encoder $\mathcal{E}_{I}(\cdot)$ and two private encoders $\mathcal{E}_{S}^1(\cdot)$ and $\mathcal{E}_{S}^2(\cdot)$ to extract the disentangled representation. Formally, $\mathcal{R}_I^1 = \mathcal{E}_{I}(X^1)$, $\mathcal{R}_I^2 = \mathcal{E}_{I}(X^2)$, $\mathcal{R}_S^1 = \mathcal{E}_{S}^{1}(X^1)$, and $\mathcal{R}_S^2 = \mathcal{E}_{S}^{2}(X^2)$. The complement extractor is a separate set of encoders with the same structure as the feature extractor. All encoders, the main predictor $F_{P}$, and auxiliary predictors ($F_{I}^{1}$, $F_{I}^{2}$, $F_{S}^{1}$, and $F_{S}^{2}$) are implemented as MLP networks with hidden layers of sizes [64, 32]. We use binary cross entropy for $\mathcal{L}_{p}$ and define $\mathcal{L}_{c}(Y, \hat{Y}) = 1/(\theta + |Y-\hat{Y}|)$, where $\theta = 0.01$ prevents division by zero. This $\mathcal{L}_{c}$ increases as the predicted label $\hat{Y}$ approaches the true label $Y$. Here, $\mathcal{L}_{task}$ in \cref{eq:pns_loss} is $\mathcal{L}_{p}(Y, F_{P}([\mathcal{R}_I^1, \mathcal{R}_I^2, \mathcal{R}_S^1, \mathcal{R}_S^2]))$.

\begin{table}[t]
\caption{Distance Correlation based on $s$ for modality 1}
\label{tab:syn-dataset-results-1}
\centering
\small
\begin{tabular}{llllll}
\hline
\multicolumn{1}{l}{} & Mode & \textbf{SN} & SF & NC & SC \\
\hline
\multirow{3}{*}{s = 0.0} 
& Net & 0.600 & 0.647 & 0.635 & 0.269 \\
& Net+MPNS(-c) & 0.608 & 0.652 & 0.545 & 0.261 \\
& Net+MPNS & \textbf{0.658} & 0.638 & 0.556 & 0.273 \\
\hline
\multirow{3}{*}{s = 0.1}
& Net & 0.590 & 0.647 & 0.640 & 0.282 \\
& Net+MPNS(-c) & 0.594 & 0.655 & 0.557 & 0.280 \\
& Net+MPNS & \textbf{0.675} & 0.613 & 0.565 & 0.285 \\
\hline
\multirow{3}{*}{s = 0.3} 
& Net & 0.591 & 0.656 & 0.617 & 0.302 \\
& Net+MPNS(-c) & 0.600 & 0.657 & 0.555 & 0.298 \\
& Net+MPNS & \textbf{0.631} & 0.634 & 0.551 & 0.302 \\
\hline
\multirow{3}{*}{s = 0.5} 
& Net & 0.593 & 0.662 & 0.625 & 0.327 \\
& Net+MPNS(-c) & 0.603 & 0.663 & 0.554 & 0.333 \\
& Net+MPNS & \textbf{0.650} & 0.648 & 0.564 & 0.342 \\
\hline
\multirow{3}{*}{s = 0.7} 
& Net & 0.594 & 0.653 & 0.640 & 0.326 \\
& Net+MPNS(-c) & 0.610 & 0.653 & 0.562 & 0.327 \\
& Net+MPNS & \textbf{0.651} & 0.632 & 0.563 & 0.338 \\
\hline
\end{tabular}
\end{table}

\begin{table}[t]
\caption{Distance Correlation based on $s$ for modality 2}
\label{tab:syn-dataset-results-2}
\centering
\small
\begin{tabular}{llllll}
\hline
\multicolumn{1}{l}{} & Mode & \textbf{SN} & SF & NC & SC \\
\hline
\multirow{3}{*}{s = 0.0}
& Net & 0.492 & 0.580 & 0.617 & 0.291 \\
& Net+MPNS(-c) & 0.563 & 0.537 & 0.592 & 0.299 \\
& Net+MPNS & \textbf{0.628} & 0.548 & 0.607 & 0.343 \\
\hline
\multirow{3}{*}{s = 0.1} 
& Net & 0.487 & 0.579 & 0.608 & 0.297 \\
& Net+MPNS(-c) & 0.543 & 0.546 & 0.573 & 0.338 \\
& Net+MPNS & \textbf{0.629} & 0.531 & 0.603 & 0.339 \\
\hline
\multirow{3}{*}{s = 0.3} 
& Net & 0.492 & 0.591 & 0.591 & 0.325 \\
& Net+MPNS(-c) & 0.564 & 0.546 & 0.584 & 0.359 \\
& Net+MPNS & \textbf{0.612} & 0.538 & 0.589 & 0.367 \\
\hline
\multirow{3}{*}{s = 0.5} 
& Net & 0.472 & 0.596 & 0.603 & 0.335 \\
& Net+MPNS(-c) & 0.540 & 0.555 & 0.585 & 0.400 \\
& Net+MPNS & \textbf{0.601} & 0.545 & 0.602 & 0.388 \\
\hline
\multirow{3}{*}{s = 0.7} 
& Net & 0.475 & 0.588 & 0.607 & 0.345 \\
& Net+MPNS(-c) & 0.562 & 0.549 & 0.585 & 0.416 \\
& Net+MPNS & \textbf{0.626} & 0.526 & 0.578 & 0.425 \\
\hline
\end{tabular}
\end{table}

\subsubsection{Implementation.}
The decomposition model (denoted as Net) is trained by optimizing only $\mathcal{L}^{task}$ in \cref{eq:pns_loss}, while its MPNS adaption (denoted as Net+MPNS) is trained by optimizing $\mathcal{L}^{total}$ in \cref{eq:pns_loss}. To evaluate their performance, for modality 1, we compute the distance correlation between the extracted representation $[\mathcal{R}_I^1, \mathcal{R}_S^1]$ and each variable type (SN, SF, NC, and SC) in $X^1$. Similarly, for modality 2, we use $[\mathcal{R}_I^2, \mathcal{R}_S^2]$ and variables in $X^2$. To evaluate the impact of constraint terms in \cref{eq:pns_I_loss} and \cref{eq:pns_S_loss}, we train a variant of Net+MPNS (denoted as Net+MPNS(-c)) by eliminating the $\mathcal{L}_{m,I}^{constr}$ and $\mathcal{L}_{m,S}^{constr}$ terms. % We run the experiment 10 times.

\subsubsection{Results and Discussion.}
Table \ref{tab:syn-dataset-results-1} and \cref{tab:syn-dataset-results-2} present the distance correlation values between the learned representations and the ground truth variables under varying degrees of spurious correlation ($s$).

Our analysis focuses on SN variables, which directly determine $Y$. A higher distance correlation indicates a better representation. Both tables demonstrate that NET+MPNS consistently outperforms both Net and NET+MPNS(-c) in capturing the SN causes across various degrees of $s$ for both modalities. This demonstrates MPNS's effectiveness in learning representations with high PNS. Also, this underscores the importance of the full optimization objective, including the constraint term, in enforcing the learning of non-trivial PNS.

Additionally, the distance correlation with spurious information increases proportionally with $s$. While MPNS captures some spurious information when data contains stronger spurious correlations, it maintains effective extraction of SN causes, demonstrating its robustness.

\begin{table}[t]
\caption{Comparison on CMU-MOSI dataset.}
\label{tab:mosi_result}
\centering
% \small
\begin{tabular}{lllll}
\hline
\multicolumn{4}{c}{\textbf{Aligned}} \\
\hline
\textbf{Methods}           & \textbf{Acc\_7(\%)} & \textbf{Acc\_2(\%)} & \textbf{F1(\%)} \\
\hline
TFN             & 32.1                & 73.9                & 73.4            \\
LMF               & 32.8                & 76.4                & 75.7            \\
MFM            & 36.2                & 78.1                & 78.1            \\
RAVEN            & 33.2                & 78.0                & 76.6            \\
MCTN             & 35.6                & 79.3                & 79.1            \\
DMD               & 35.9                & 79.0                & 79.0             \\
DMD+MPNS(-c)      & 35.1                & 79.6                & 79.3             \\
DMD+MPNS   & \textbf{36.4}       & \textbf{79.8}       & \textbf{79.8}    \\
\hline
\multicolumn{4}{c}{\textbf{Unaligned}} \\
\hline
\textbf{Methods}           & \textbf{Acc\_7(\%)} & \textbf{Acc\_2(\%)} & \textbf{F1(\%)} \\
\hline
% \textbf{Methods}           & \textbf{Acc\_7(\%)} & \textbf{Acc\_2(\%)} & \textbf{F1(\%)} \\
% \hline
RAVEN            &  31.7                & 72.7                & 73.1            \\
MCTN             & 32.7                & 75.9                & 76.4            \\
DMD              & 35.9                & 78.8                & 78.9            \\
DMD+MPNS(-c)     & 35.4                & 79.3                & 79.4            \\
DMD+MPNS   & \textbf{36.3}       & \textbf{79.7}       & \textbf{79.7}    \\
\hline
\end{tabular}
\end{table}

\begin{table}[t]
\caption{Comparison on CMU-MOSEI dataset.}
\label{tab:mosei_result}
\centering
% \small
\begin{tabular}{lllll}
\hline
\multicolumn{4}{c}{\textbf{Aligned}} \\
\hline
\textbf{Methods}  & \textbf{Acc\_7(\%)} & \textbf{Acc\_2(\%)} & \textbf{F1(\%)} \\
\hline
Graph-MFN        & 45.0                & 76.9                & 77.0            \\
RAVEN             & 50.0                & 79.1                & 79.5            \\
MCTN            & 49.6                & 79.8                & 80.6            \\
DMD               & 51.8                & 83.8                & 83.3            \\
DMD+MPNS(-c)     & 52.0                & 83.3                & 83.4            \\
DMD+MPNS   & \textbf{52.2}       & \textbf{84.4}       & \textbf{84.2}   \\
\hline
\multicolumn{4}{c}{\textbf{Unaligned}} \\
\hline
\textbf{Methods}  & \textbf{Acc\_7(\%)} & \textbf{Acc\_2(\%)} & \textbf{F1(\%)} \\
\hline
% \textbf{Methods}  & \textbf{Acc\_7(\%)} & \textbf{Acc\_2(\%)} & \textbf{F1(\%)} \\
% \hline
RAVEN            & 45.5                & 75.4                & 75.7            \\
MCTN               & 48.2                & 79.3                & 79.7            \\
DMD              & 52.0                & 83.2                & 83.1            \\
DMD+MPNS(-c)      & 52.3                & 84.1                & 84.0            \\
DMD+MPNS   & \textbf{53.2}       & \textbf{84.4}       & \textbf{84.2}  \\
\hline
\end{tabular}
\end{table}

\subsection{Real-world Dataset Experiments}
We conduct extensive experiments to demonstrate that MPNS can improve both the predictive performance and robustness of multimodal learning. Specifically, we evaluate our method on standard multimodal prediction tasks and under modality-missing scenarios.

\subsubsection{Real-world Datasets.}
We utilize CMU-MOSI \cite{zadeh2016multimodal} and CMU-MOSEI \cite{zadeh2018multimodal}, two widely-used datasets for multimodal emotion recognition. Both datasets contain three modalities: language ($l$), vision ($v$), and acoustic ($a$), and provide samples labeled with sentiment scores ranging from highly negative (-3) to highly positive (3).
CMU-MOSI consists of 2,199 short monologue video clips, split into 1,284 training, 229 validation, and 686 testing samples. CMU-MOSEI, a larger dataset, contains 22,856 movie review video clips from YouTube, divided into 16,326 training, 1,871 validation, and 4,659 testing samples.

\subsubsection{Base decomposition model.}
We implement MPNS by adapting the Decoupled Multimodal Distillation (DMD) \cite{li2023decoupled}, a state-of-the-art decomposition model. For its feature extractor, DMD uses a shared multimodal encoder to extract modality-invariant representations and private encoders for modality-specific representations from multimodal data. It also employs knowledge distillation to improve feature extraction, followed by a main predictor and auxiliary predictors for outcome prediction.

\subsubsection{Implementation.}
To implement MPNS, we utilize the DMD and its hyper-parameters based on its publicly available code\footnote{https://github.com/mdswyz/DMD}. We then add a complement extractor mirroring the architecture of DMD's feature extractor. To optimize \cref{eq:pns_loss}, we empirically define $\mathcal{L}_{p}$ as the mean absolute error (MAE) and $\mathcal{L}_{c}(Y, \hat{Y}) = max(0, 4 - ||MAE(Y, \hat{Y})||)$. This $\mathcal{L}_{c}$ increases as the predicted label $\hat{Y}$ approaches the true $Y$. $\mathcal{L}^{task}$ is the original DMD loss. By adapting DMD according to \cref{fig:method}, we create DMD+MPNS, which optimizes the full $\mathcal{L}^{total}$ in \cref{eq:pns_loss}. To evaluate the impact of the constraint terms in \cref{eq:pns_I_loss} and \cref{eq:pns_S_loss}, we train DMD+MPNS(-c), a variant that eliminates the $\mathcal{L}_{m,I}^{constr}$ and $\mathcal{L}_{m,S}^{constr}$ terms.

To assess our method's impact on model performance, we evaluate DMD, DMD+MPNS(-c), and DMD+MPNS while comparing them with state-of-the-art methods for emotion score prediction under the same dataset settings: TFN \cite{zadeh2017tensor}, LMF \cite{liu2018efficient}, MFM \cite{tsai2019learning}, RAVEN \cite{wang2019words}, MCTN \cite{pham2019found}, and Graph-MFN \cite{zadeh2018multimodal}. Following these works, we evaluate the performance using: (1) 7-class accuracy (Acc\_7), (2) binary accuracy (Acc\_2), and (3) F1 score (F1).

To investigate whether MPNS can enhance model robustness under missing modalities, we conduct additional experiments with modality dropout during training and testing. During training, we randomly drop 0, 1, or 2 modalities with equal probability for each input sample. During testing, we evaluate models under fixed modality-missing scenarios by systematically removing different combinations of modalities.

\begin{table}[t]
\caption{Performance Metrics (Acc\_7(\%) /Acc\_2 (\%)/F1 (\%)) on CMU-MOSI with Missing Modalities.}
\label{tab:mosi_result_miss}
\centering
\footnotesize
\begin{tabular}{lllll}
\hline
\multicolumn{4}{c}{\textbf{Aligned}} \\
\hline
\makecell[l]{\textbf{Missing} \\ \textbf{Modality}} & \textbf{DMD} & \makecell[l]{\textbf{DMD+} \\ \textbf{MPNS(-c)}} & \makecell[l]{\textbf{DMD+} \\ \textbf{MPNS}} \\
\hline
$\{l\}$             &     16.9 / 46.5 / 40.0            & 17.3 / 47.2 / 40.2                & \textbf{18.3} / \textbf{48.4} / \textbf{40.9}            \\
$\{a\}$               & 34.4 / 77.7 / 77.9                & 34.2 / 78.1 / 77.6                & \textbf{34.6} / \textbf{78.3} / \textbf{78.3}            \\
$\{v\}$            & 33.8 / 78.3 / 78.4                & 34.2 / 77.9 / \textbf{78.5}                & \textbf{34.7} / \textbf{78.9} / 78.4            \\
$\{l,a\}$   &        14.2 / 43.8 / \textbf{38.2}            & 15.1 / 43.8 / 38.1                & \textbf{16.5} / \textbf{45.3} / 37.9            \\
$\{l,v\}$             & 15.1 / 44.7 / 39.4            & 15.1 / 44.6 / 39.0                & \textbf{16.5} / \textbf{45.3} / \textbf{40.1}            \\
$\{a,v\}$               & 33.1 / 77.9 / 78.2                & 33.9 / 77.8 / 78.3                & \textbf{34.2} / \textbf{78.3} / \textbf{78.3}       \\
\hline
\multicolumn{4}{c}{\textbf{Unaligned}} \\
\hline
\makecell[l]{\textbf{Missing} \\ \textbf{Modality}} & \textbf{DMD} & \makecell[l]{\textbf{DMD+} \\ \textbf{MPNS(-c)}} & \makecell[l]{\textbf{DMD+} \\ \textbf{MPNS}} \\
\hline
% \makecell[l]{\textbf{Missing} \\ \textbf{Modality}} & \textbf{DMD} & \makecell[l]{\textbf{DMD+} \\ \textbf{MPNS(-c)}} & \makecell[l]{\textbf{DMD+} \\ \textbf{MPNS}} \\
% \hline
$\{l\}$             & 16.4 / 46.1 / 39.5            & 17.1 / 47.0 / 39.8                & \textbf{18.1} / \textbf{48.9} / \textbf{41.2}            \\
$\{a\}$             & 33.9 / 77.2 / 77.4            & 34.0 / 77.8 / 77.9                & \textbf{34.3} / \textbf{78.8} / \textbf{78.8}            \\
$\{v\}$             & \textbf{34.6} / 77.8 / 77.9            & 34.0 / 78.2 / 78.1                & 34.4 / \textbf{79.1} / \textbf{78.9}            \\
$\{l,a\}$           & 14.8 / 43.2 / 37.8            & 14.7 / 44.1 / 37.5                & \textbf{16.3} / \textbf{45.8} / \textbf{38.4}            \\
$\{l,v\}$           & 14.7 / 44.1 / 38.9            & 14.9 / 44.3 / 39.3                & \textbf{17.0} / \textbf{45.8} / \textbf{40.6}            \\
$\{a,v\}$           & 34.2 / 77.3 / 77.8            & 33.8 / 77.6 / 78.0                & \textbf{34.7} / \textbf{78.8} / \textbf{78.8}            \\
\hline
\end{tabular}
\end{table}

\begin{table}
\caption{Performance Metrics (Acc\_7(\%) /Acc\_2 (\%)/F1 (\%)) on CMU-MOSEI with Missing Modalities.}
\label{tab:mosei_result_miss}
\centering
\footnotesize
\begin{tabular}{lllll}
\hline

\multicolumn{4}{c}{\textbf{Aligned}} \\
\hline
\makecell[l]{\textbf{Missing} \\ \textbf{Modality}} & \textbf{DMD} & \makecell[l]{\textbf{DMD+} \\ \textbf{MPNS(-c)}} & \makecell[l]{\textbf{DMD+} \\ \textbf{MPNS}} \\
\hline
$\{l\}$             & 42.8 / \textbf{66.2} / 63.6             & 42.9 / 63.8 / 65.9                & \textbf{43.8} / 64.7 / \textbf{66.1}            \\
$\{a\}$             & 51.2 / 81.7 / 81.8                      & 50.8 / 82.3 / 81.9                & \textbf{51.4} / \textbf{82.6} / \textbf{82.5}   \\
$\{v\}$             & 50.5 / 80.3 / 82.1                      & 49.9 / 81.2 / 81.5                & \textbf{52.0} / \textbf{82.8} / \textbf{82.9}   \\
$\{l,a\}$           & 42.1 / 63.9 / 62.7                      & 41.8 / 62.8 / 65.3                & \textbf{42.6} / \textbf{65.7} / \textbf{65.8}   \\ 
$\{l,v\}$           & 41.8 / 62.8 / 63.8                      & 40.9 / 64.1 / 64.2                & \textbf{43.2} / \textbf{64.5} / \textbf{64.9}   \\
$\{a,v\}$           & 49.3 / 79.5 / 81.0                      & 48.9 / 79.9 / 80.2                & \textbf{50.9} / \textbf{81.1} / \textbf{80.2}    \\
\hline
\multicolumn{4}{c}{\textbf{Unaligned}} \\
\hline
\makecell[l]{\textbf{Missing} \\ \textbf{Modality}} & \textbf{DMD} & \makecell[l]{\textbf{DMD+} \\ \textbf{MPNS(-c)}} & \makecell[l]{\textbf{DMD+} \\ \textbf{MPNS}} \\
\hline
% \makecell[l]{\textbf{Missing} \\ \textbf{Modality}} & \textbf{DMD} & \makecell[l]{\textbf{DMD+} \\ \textbf{MPNS(-c)}} & \makecell[l]{\textbf{DMD+} \\ \textbf{MPNS}} \\
% \hline
$\{l\}$             & 41.5 / 64.8 / 64.1             & 42.1 / 65.2 / 65.0                & \textbf{43.9} / \textbf{66.5} / \textbf{66.9}            \\
$\{a\}$             & 49.8 / 80.2 / 80.5             & 50.1 / 81.0 / 81.2                & \textbf{52.3} / \textbf{82.9} / \textbf{83.1}   \\
$\{v\}$             & 49.2 / 79.1 / 80.8             & 49.5 / 80.5 / 81.0                & \textbf{51.8} / \textbf{82.5} / \textbf{83.2}   \\
$\{l,a\}$           & 40.8 / 64.5 / 63.9             & 41.2 / 63.1 / 64.2                & \textbf{43.1} / \textbf{65.8} / \textbf{66.2}   \\ 
$\{l,v\}$           & 40.5 / 63.9 / 64.5             & 41.2 / 63.2 / 63.8                & \textbf{42.9} / \textbf{64.8} / \textbf{65.2}   \\
$\{a,v\}$           & 48.1 / 78.2 / 79.5             & 48.8 / 79.1 / 79.8                & \textbf{51.2} / \textbf{81.3} / \textbf{80.9}    \\
\hline
\end{tabular}
\end{table}

\subsubsection{Results and Discussion.}
The experimental results for predictive performance on CMU-MOSI and CMU-MOSEI are presented in \cref{tab:mosi_result} and \cref{tab:mosei_result}, respectively. The results demonstrate that MPNS implementation enhances DMD's performance across all evaluation metrics on both datasets, regardless of whether the data is aligned or unaligned. This enhancement validates the effectiveness of encouraging the decomposition model to learn high-PNS representations. By focusing on features that are both necessary and sufficient for accurate predictions, the model learns more informative and discriminative representations, leading to better performance.

The results for modality missing scenarios are shown in \cref{tab:mosi_result_miss} and \cref{tab:mosei_result_miss}. DMD+MPNS outperforms both DMD and DMD+MPNS(-c) in most cases. This enhanced robustness could be attributed to MPNS's ability to learn representations that contain necessary and sufficient predictive information through its PNS optimization objective.

Furthermore, the comparative analysis of DMD+MPNS(-c) reveals that eliminating the constraint term leads to decreased performance relative to the complete DMD+MPNS model in both standard and modality missing scenarios. This highlights the importance of using constraints to ensure that the multimodal representations capture the desired high-PNS properties, contributing to both performance and robustness improvements.

\section{Limitation}
\label{sec:limitation}
MPNS builds decomposition models for learning effective representations through PNS incorporation. However, completely and successfully decomposing the representation into modality-invariant and modality-specific components is an open problem in the field \cite{zhang2019multimodal, ramachandram2017deep, guo2019deep, gao2020survey}. The process itself may introduce noise, which could affect the performance of MPNS. Despite this, we believe that MPNS offers novel insights into multimodal representation learning.

\section{Conclusion}
Our study extends PNS estimation into multimodal representation learning and proposes viewing multimodal representations as comprising modality-invariant and modality-specific components to address these challenges. Building upon the derivations of PNS for these components, we develop a method that enhances multimodal models by encouraging them to learn representations with high PNS. Experiments on synthetic and real-world datasets validate our method's effectiveness in enhancing both predictive performance and robustness of multimodal learning.

\newpage
% \input{sec/1_intro}
% \input{sec/2_formatting}
% \input{sec/3_finalcopy}

% {
%     \small
%     \bibliographystyle{ieeenat_fullname}
%     % \bibliographystyle{unsrt}
%     \bibliography{main}
% }


\end{document}

%% file: main.bbl
\begin{thebibliography}{50}
\providecommand{\natexlab}[1]{#1}
\providecommand{\url}[1]{\texttt{#1}}
\expandafter\ifx\csname urlstyle\endcsname\relax
  \providecommand{\doi}[1]{doi: #1}\else
  \providecommand{\doi}{doi: \begingroup \urlstyle{rm}\Url}\fi

\bibitem[Ahuja et~al.(2020)Ahuja, Shanmugam, Varshney, and Dhurandhar]{ahuja2020invariant}
Kartik Ahuja, Karthikeyan Shanmugam, Kush Varshney, and Amit Dhurandhar.
\newblock Invariant risk minimization games.
\newblock In \emph{International Conference on Machine Learning}, pages 145--155. PMLR, 2020.

\bibitem[Arjovsky et~al.(2019)Arjovsky, Bottou, Gulrajani, and Lopez-Paz]{arjovsky2019invariant}
Martin Arjovsky, L{\'e}on Bottou, Ishaan Gulrajani, and David Lopez-Paz.
\newblock Invariant risk minimization.
\newblock \emph{arXiv preprint arXiv:1907.02893}, 2019.

\bibitem[Cai et~al.(2024)Cai, Wang, Jordan, and Song]{cai2024learning}
Hengrui Cai, Yixin Wang, Michael Jordan, and Rui Song.
\newblock On learning necessary and sufficient causal graphs.
\newblock \emph{Advances in Neural Information Processing Systems}, 36, 2024.

\bibitem[Chen et~al.(2024{\natexlab{a}})Chen, Solebo, Bao, and Taylor]{chen2024medical}
Boyu Chen, Ameenat~L Solebo, Weiye Bao, and Paul Taylor.
\newblock Medical image quality assessment based on probability of necessity and sufficiency.
\newblock \emph{arXiv preprint arXiv:2410.08118}, 2024{\natexlab{a}}.

\bibitem[Chen et~al.(2024{\natexlab{b}})Chen, Cai, Zheng, Jiang, Huang, Hao, and Li]{chen2024unifying}
Xuexin Chen, Ruichu Cai, Kaitao Zheng, Zhifan Jiang, Zhengting Huang, Zhifeng Hao, and Zijian Li.
\newblock Unifying invariance and spuriousity for graph out-of-distribution via probability of necessity and sufficiency.
\newblock \emph{arXiv preprint arXiv:2402.09165}, 2024{\natexlab{b}}.

\bibitem[Chen et~al.(2022)Chen, Zhang, Bian, Yang, Kaili, Xie, Liu, Han, and Cheng]{chen2022learning}
Yongqiang Chen, Yonggang Zhang, Yatao Bian, Han Yang, MA Kaili, Binghui Xie, Tongliang Liu, Bo Han, and James Cheng.
\newblock Learning causally invariant representations for out-of-distribution generalization on graphs.
\newblock \emph{Advances in Neural Information Processing Systems}, 35:\penalty0 22131--22148, 2022.

\bibitem[Dong et~al.(2023)Dong, Han, Peng, Qi, Ge, Yang, Zhao, Sun, Zhou, Wei, et~al.]{dong2023dreamllm}
Runpei Dong, Chunrui Han, Yuang Peng, Zekun Qi, Zheng Ge, Jinrong Yang, Liang Zhao, Jianjian Sun, Hongyu Zhou, Haoran Wei, et~al.
\newblock Dreamllm: Synergistic multimodal comprehension and creation.
\newblock In \emph{The Twelfth International Conference on Learning Representations}, 2023.

\bibitem[Gamella and Heinze-Deml(2020)]{gamella2020active}
Juan~L Gamella and Christina Heinze-Deml.
\newblock Active invariant causal prediction: Experiment selection through stability.
\newblock \emph{Advances in Neural Information Processing Systems}, 33:\penalty0 15464--15475, 2020.

\bibitem[Gao et~al.(2020)Gao, Li, Chen, and Zhang]{gao2020survey}
Jing Gao, Peng Li, Zhikui Chen, and Jianing Zhang.
\newblock A survey on deep learning for multimodal data fusion.
\newblock \emph{Neural Computation}, 32\penalty0 (5):\penalty0 829--864, 2020.

\bibitem[Guo et~al.(2019)Guo, Wang, and Wang]{guo2019deep}
Wenzhong Guo, Jianwen Wang, and Shiping Wang.
\newblock Deep multimodal representation learning: A survey.
\newblock \emph{Ieee Access}, 7:\penalty0 63373--63394, 2019.

\bibitem[Hu et~al.(2018)Hu, Chen, Partovi~Nia, Chan, and Geng]{hu2018causal}
Shoubo Hu, Zhitang Chen, Vahid Partovi~Nia, Laiwan Chan, and Yanhui Geng.
\newblock Causal inference and mechanism clustering of a mixture of additive noise models.
\newblock \emph{Advances in neural information processing systems}, 31, 2018.

\bibitem[Huang et~al.(2020)Huang, Zhang, Zhang, Ramsey, Sanchez-Romero, Glymour, and Sch{\"o}lkopf]{huang2020causal}
Biwei Huang, Kun Zhang, Jiji Zhang, Joseph Ramsey, Ruben Sanchez-Romero, Clark Glymour, and Bernhard Sch{\"o}lkopf.
\newblock Causal discovery from heterogeneous/nonstationary data.
\newblock \emph{Journal of Machine Learning Research}, 21\penalty0 (89):\penalty0 1--53, 2020.

\bibitem[Islam et~al.(2023)Islam, Gladstone, Islam, and Iqbal]{islam2023eqa}
Md~Mofijul Islam, Alexi Gladstone, Riashat Islam, and Tariq Iqbal.
\newblock Eqa-mx: Embodied question answering using multimodal expression.
\newblock In \emph{The Twelfth International Conference on Learning Representations}, 2023.

\bibitem[Jones et~al.(1995)Jones, Forrest, et~al.]{jones1995fitness}
Terry Jones, Stephanie Forrest, et~al.
\newblock Fitness distance correlation as a measure of problem difficulty for genetic algorithms.
\newblock In \emph{ICGA}, pages 184--192, 1995.

\bibitem[Li et~al.(2023)Li, Wang, and Cui]{li2023decoupled}
Yong Li, Yuanzhi Wang, and Zhen Cui.
\newblock Decoupled multimodal distilling for emotion recognition.
\newblock In \emph{Proceedings of the IEEE/CVF Conference on Computer Vision and Pattern Recognition}, pages 6631--6640, 2023.

\bibitem[Liang et~al.(2024{\natexlab{a}})Liang, Yu, Yang, Brown, Cui, Zhao, Gong, and Zhou]{liang2024module}
Chen Liang, Jiahui Yu, Ming-Hsuan Yang, Matthew Brown, Yin Cui, Tuo Zhao, Boqing Gong, and Tianyi Zhou.
\newblock Module-wise adaptive distillation for multimodality foundation models.
\newblock \emph{Advances in Neural Information Processing Systems}, 36, 2024{\natexlab{a}}.

\bibitem[Liang et~al.(2024{\natexlab{b}})Liang, Cheng, Fan, Ling, Nie, Chen, Deng, Allen, Auerbach, Mahmood, et~al.]{liang2024quantifying}
Paul~Pu Liang, Yun Cheng, Xiang Fan, Chun~Kai Ling, Suzanne Nie, Richard Chen, Zihao Deng, Nicholas Allen, Randy Auerbach, Faisal Mahmood, et~al.
\newblock Quantifying \& modeling multimodal interactions: An information decomposition framework.
\newblock \emph{Advances in Neural Information Processing Systems}, 36, 2024{\natexlab{b}}.

\bibitem[Liu et~al.(2021)Liu, Sun, Wang, Tang, Li, Qin, Chen, and Liu]{liu2021learning}
Chang Liu, Xinwei Sun, Jindong Wang, Haoyue Tang, Tao Li, Tao Qin, Wei Chen, and Tie-Yan Liu.
\newblock Learning causal semantic representation for out-of-distribution prediction.
\newblock \emph{Advances in Neural Information Processing Systems}, 34:\penalty0 6155--6170, 2021.

\bibitem[Liu et~al.(2024{\natexlab{a}})Liu, Kimura, Liu, Wang, Li, Diggavi, Srivastava, and Abdelzaher]{liu2024focal}
Shengzhong Liu, Tomoyoshi Kimura, Dongxin Liu, Ruijie Wang, Jinyang Li, Suhas Diggavi, Mani Srivastava, and Tarek Abdelzaher.
\newblock Focal: Contrastive learning for multimodal time-series sensing signals in factorized orthogonal latent space.
\newblock \emph{Advances in Neural Information Processing Systems}, 36, 2024{\natexlab{a}}.

\bibitem[Liu et~al.(2024{\natexlab{b}})Liu, Wang, Ying, and Zhao]{liu2024muse}
Tianyu Liu, Yuge Wang, Rex Ying, and Hongyu Zhao.
\newblock Muse-gnn: Learning unified gene representation from multimodal biological graph data.
\newblock \emph{Advances in Neural Information Processing Systems}, 36, 2024{\natexlab{b}}.

\bibitem[Liu and Shen(2018)]{liu2018efficient}
Zhun Liu and Ying Shen.
\newblock Efficient low-rank multimodal fusion with modality-specific factors.
\newblock In \emph{Proceedings of the 56th Annual Meeting of the Association for Computational Linguistics (Long Papers)}, 2018.

\bibitem[Locatello et~al.(2019)Locatello, Bauer, Lucic, Raetsch, Gelly, Sch{\"o}lkopf, and Bachem]{locatello2019challenging}
Francesco Locatello, Stefan Bauer, Mario Lucic, Gunnar Raetsch, Sylvain Gelly, Bernhard Sch{\"o}lkopf, and Olivier Bachem.
\newblock Challenging common assumptions in the unsupervised learning of disentangled representations.
\newblock In \emph{international conference on machine learning}, pages 4114--4124. PMLR, 2019.

\bibitem[Louizos et~al.(2017)Louizos, Shalit, Mooij, Sontag, Zemel, and Welling]{louizos2017causal}
Christos Louizos, Uri Shalit, Joris~M Mooij, David Sontag, Richard Zemel, and Max Welling.
\newblock Causal effect inference with deep latent-variable models.
\newblock \emph{Advances in neural information processing systems}, 30, 2017.

\bibitem[Lu et~al.(2021)Lu, Wu, Hern{\'a}ndez-Lobato, and Sch{\"o}lkopf]{lu2021invariant}
Chaochao Lu, Yuhuai Wu, Jos{\'e}~Miguel Hern{\'a}ndez-Lobato, and Bernhard Sch{\"o}lkopf.
\newblock Invariant causal representation learning for out-of-distribution generalization.
\newblock In \emph{International Conference on Learning Representations}, 2021.

\bibitem[Mai et~al.(2020)Mai, Hu, and Xing]{mai2020modality}
Sijie Mai, Haifeng Hu, and Songlong Xing.
\newblock Modality to modality translation: An adversarial representation learning and graph fusion network for multimodal fusion.
\newblock In \emph{Proceedings of the AAAI Conference on Artificial Intelligence}, pages 164--172, 2020.

\bibitem[Pearl(2009)]{pearl2009causality}
Judea Pearl.
\newblock \emph{Causality}.
\newblock Cambridge university press, 2009.

\bibitem[Peters et~al.(2014)Peters, Mooij, Janzing, and Sch{\"o}lkopf]{peters2014causal}
Jonas Peters, Joris~M Mooij, Dominik Janzing, and Bernhard Sch{\"o}lkopf.
\newblock Causal discovery with continuous additive noise models.
\newblock \emph{Journal of Machine Learning Research}, 15\penalty0 (58):\penalty0 2009--2053, 2014.

\bibitem[Pham et~al.(2019)Pham, Liang, Manzini, Morency, and P{\'o}czos]{pham2019found}
Hai Pham, Paul~Pu Liang, Thomas Manzini, Louis-Philippe Morency, and Barnab{\'a}s P{\'o}czos.
\newblock Found in translation: Learning robust joint representations by cyclic translations between modalities.
\newblock In \emph{Proceedings of the AAAI conference on artificial intelligence}, pages 6892--6899, 2019.

\bibitem[Ramachandram and Taylor(2017)]{ramachandram2017deep}
Dhanesh Ramachandram and Graham~W Taylor.
\newblock Deep multimodal learning: A survey on recent advances and trends.
\newblock \emph{IEEE signal processing magazine}, 34\penalty0 (6):\penalty0 96--108, 2017.

\bibitem[Sch{\"o}lkopf et~al.(2021)Sch{\"o}lkopf, Locatello, Bauer, Ke, Kalchbrenner, Goyal, and Bengio]{scholkopf2021toward}
Bernhard Sch{\"o}lkopf, Francesco Locatello, Stefan Bauer, Nan~Rosemary Ke, Nal Kalchbrenner, Anirudh Goyal, and Yoshua Bengio.
\newblock Toward causal representation learning.
\newblock \emph{Proceedings of the IEEE}, 109\penalty0 (5):\penalty0 612--634, 2021.

\bibitem[Shi et~al.(2019)Shi, Paige, Torr, et~al.]{shi2019variational}
Yuge Shi, Brooks Paige, Philip Torr, et~al.
\newblock Variational mixture-of-experts autoencoders for multi-modal deep generative models.
\newblock \emph{Advances in neural information processing systems}, 32, 2019.

\bibitem[Swamy et~al.(2024)Swamy, Satayeva, Frej, Bossy, Vogels, Jaggi, K{\"a}ser, and Hartley]{swamy2024multimodn}
Vinitra Swamy, Malika Satayeva, Jibril Frej, Thierry Bossy, Thijs Vogels, Martin Jaggi, Tanja K{\"a}ser, and Mary-Anne Hartley.
\newblock Multimodn—multimodal, multi-task, interpretable modular networks.
\newblock \emph{Advances in Neural Information Processing Systems}, 36, 2024.

\bibitem[Tang et~al.(2024)Tang, Du, Vo, Lal, and Huth]{tang2024brain}
Jerry Tang, Meng Du, Vy Vo, Vasudev Lal, and Alexander Huth.
\newblock Brain encoding models based on multimodal transformers can transfer across language and vision.
\newblock \emph{Advances in Neural Information Processing Systems}, 36, 2024.

\bibitem[Tran et~al.(2024)Tran, Dicente~Cid, Lahiani, Theis, Peng, and Klaiman]{tran2024training}
Manuel Tran, Yashin Dicente~Cid, Amal Lahiani, Fabian Theis, Tingying Peng, and Eldad Klaiman.
\newblock Training transitive and commutative multimodal transformers with loretta.
\newblock \emph{Advances in Neural Information Processing Systems}, 36, 2024.

\bibitem[Tsai et~al.(2018)Tsai, Liang, Zadeh, Morency, and Salakhutdinov]{tsai2018learning}
Yao-Hung~Hubert Tsai, Paul~Pu Liang, Amir Zadeh, Louis-Philippe Morency, and Ruslan Salakhutdinov.
\newblock Learning factorized multimodal representations.
\newblock In \emph{International Conference on Learning Representations}, 2018.

\bibitem[Tsai et~al.(2019)Tsai, Liang, Zadeh, Morency, and Salakhutdinov]{tsai2019learning}
Yao-Hung~Hubert Tsai, Paul~Pu Liang, Amir Zadeh, Louis-Philippe Morency, and Ruslan Salakhutdinov.
\newblock Learning factorized multimodal representations.
\newblock In \emph{International Conference on Representation Learning}, 2019.

\bibitem[Wang et~al.(2017)Wang, Yang, Xu, Hanjalic, and Shen]{wang2017adversarial}
Bokun Wang, Yang Yang, Xing Xu, Alan Hanjalic, and Heng~Tao Shen.
\newblock Adversarial cross-modal retrieval.
\newblock In \emph{Proceedings of the 25th ACM international conference on Multimedia}, pages 154--162, 2017.

\bibitem[Wang and Jordan(2021)]{wang2021desiderata}
Yixin Wang and Michael~I Jordan.
\newblock Desiderata for representation learning: A causal perspective.
\newblock \emph{arXiv preprint arXiv:2109.03795}, 2021.

\bibitem[Wang et~al.(2019)Wang, Shen, Liu, Liang, Zadeh, and Morency]{wang2019words}
Yansen Wang, Ying Shen, Zhun Liu, Paul~Pu Liang, Amir Zadeh, and Louis-Philippe Morency.
\newblock Words can shift: Dynamically adjusting word representations using nonverbal behaviors.
\newblock In \emph{Proceedings of the AAAI Conference on Artificial Intelligence}, pages 7216--7223, 2019.

\bibitem[Xu et~al.(2024)Xu, Hou, Niell, and Beyeler]{xu2024multimodal}
Aiwen Xu, Yuchen Hou, Cristopher Niell, and Michael Beyeler.
\newblock Multimodal deep learning model unveils behavioral dynamics of v1 activity in freely moving mice.
\newblock \emph{Advances in Neural Information Processing Systems}, 36, 2024.

\bibitem[Yang et~al.(2024)Yang, Zhang, Fang, Du, Liu, Ton, Wang, and Wang]{yang2024invariant}
Mengyue Yang, Yonggang Zhang, Zhen Fang, Yali Du, Furui Liu, Jean-Francois Ton, Jianhong Wang, and Jun Wang.
\newblock Invariant learning via probability of sufficient and necessary causes.
\newblock \emph{Advances in Neural Information Processing Systems}, 36, 2024.

\bibitem[Yang et~al.(2021)Yang, Yu, Cao, Liu, Wang, and Li]{yang2021learning}
Shuai Yang, Kui Yu, Fuyuan Cao, Lin Liu, Hao Wang, and Jiuyong Li.
\newblock Learning causal representations for robust domain adaptation.
\newblock \emph{IEEE Transactions on Knowledge and Data Engineering}, 35\penalty0 (3):\penalty0 2750--2764, 2021.

\bibitem[Zadeh et~al.(2016)Zadeh, Zellers, Pincus, and Morency]{zadeh2016multimodal}
Amir Zadeh, Rowan Zellers, Eli Pincus, and Louis-Philippe Morency.
\newblock Multimodal sentiment intensity analysis in videos: Facial gestures and verbal messages.
\newblock \emph{IEEE Intelligent Systems}, 31\penalty0 (6):\penalty0 82--88, 2016.

\bibitem[Zadeh et~al.(2017)Zadeh, Chen, Poria, Cambria, and Morency]{zadeh2017tensor}
Amir Zadeh, Minghai Chen, Soujanya Poria, Erik Cambria, and Louis-Philippe Morency.
\newblock Tensor fusion network for multimodal sentiment analysis.
\newblock In \emph{Proceedings of the 2017 Conference on Empirical Methods in Natural Language Processing}. Association for Computational Linguistics, 2017.

\bibitem[Zadeh et~al.(2018)Zadeh, Liang, Poria, Cambria, and Morency]{zadeh2018multimodal}
AmirAli~Bagher Zadeh, Paul~Pu Liang, Soujanya Poria, Erik Cambria, and Louis-Philippe Morency.
\newblock Multimodal language analysis in the wild: Cmu-mosei dataset and interpretable dynamic fusion graph.
\newblock In \emph{Proceedings of the 56th Annual Meeting of the Association for Computational Linguistics (Volume 1: Long Papers)}, pages 2236--2246, 2018.

\bibitem[Zhang et~al.(2020)Zhang, Lyle, Sodhani, Filos, Kwiatkowska, Pineau, Gal, and Precup]{zhang2020invariant}
Amy Zhang, Clare Lyle, Shagun Sodhani, Angelos Filos, Marta Kwiatkowska, Joelle Pineau, Yarin Gal, and Doina Precup.
\newblock Invariant causal prediction for block mdps.
\newblock In \emph{International Conference on Machine Learning}, pages 11214--11224. PMLR, 2020.

\bibitem[Zhang et~al.(2019)Zhang, Zhai, Xie, Zhan, and Wang]{zhang2019multimodal}
Su-Fang Zhang, Jun-Hai Zhai, Bo-Jun Xie, Yan Zhan, and Xin Wang.
\newblock Multimodal representation learning: Advances, trends and challenges.
\newblock In \emph{2019 International Conference on Machine Learning and Cybernetics (ICMLC)}, pages 1--6. IEEE, 2019.

\bibitem[Zhang et~al.(2023)Zhang, Kong, and Zhou]{zhang2023integration}
Yixuan Zhang, Quyu Kong, and Feng Zhou.
\newblock Integration-free training for spatio-temporal multimodal covariate deep kernel point processes.
\newblock \emph{Advances in Neural Information Processing Systems}, 36:\penalty0 25031--25049, 2023.

\bibitem[Zheng et~al.(2018)Zheng, Aragam, Ravikumar, and Xing]{zheng2018dags}
Xun Zheng, Bryon Aragam, Pradeep~K Ravikumar, and Eric~P Xing.
\newblock Dags with no tears: Continuous optimization for structure learning.
\newblock \emph{Advances in neural information processing systems}, 31, 2018.

\bibitem[Zhu et~al.(2019)Zhu, Ng, and Chen]{zhu2019causal}
Shengyu Zhu, Ignavier Ng, and Zhitang Chen.
\newblock Causal discovery with reinforcement learning.
\newblock \emph{arXiv preprint arXiv:1906.04477}, 2019.

\end{thebibliography}
